\documentclass[letterpaper, 10 pt, conference]{ieeeconf}  

\IEEEoverridecommandlockouts                              

\usepackage{graphicx}
\usepackage{blindtext}
\usepackage{gensymb}
\usepackage{flushend}
\usepackage{pdfpages}
\usepackage{hyperref}

  \newcommand{\bu}{\mathbf{u}}
  \newcommand{\bv}{\mathbf{v}}
  \newcommand{\bo}{\mathbf{o}}
  \newcommand{\ba}{\mathbf{a}}

  \newcommand{\bs}{\mathbf{s}}

  \newcommand{\bS}{\mathbf{S}}
  \newcommand{\etal}{{\em et al.}}
  \newcommand{\ourdataset}{ROBOTPUSH }

  \title{\LARGE \bf Self-supervised Transfer Learning for Instance Segmentation through Physical Interaction }

  \author{Andreas Eitel \and Nico Hauff \and Wolfram
    Burgard
    \thanks{All authors are with the laboratory for Autonomous
      Intelligent Systems, University of Freiburg, Germany. Wolfram
      Burgard is also with the Toyota Research Institute, Los Altos,
      USA. }%
}

\begin{document}

\onecolumn

\textcopyright 2019 IEEE.  Personal use of this material is permitted.  Permission from IEEE must be obtained for all other uses, in any current or future media, including reprinting/republishing this material for advertising or promotional purposes, creating new collective works, for resale or redistribution to servers or lists, or reuse of any copyrighted component of this work in other works.
\newline
\newline
This is an extended version of the article that appeared at IEEE/RSJ International Conference on Intelligent Robots and Systems (IROS), Macau, China, 2019.
\newline
\newline
Please cite this paper as:\\
A. Eitel, N. Hauff, and W. Burgard, ``Self-supervised Transfer Learning for Instance Segmentation through Physical Interaction,'' in 2019 IEEE/RSJ International Conference on Intelligent Robots and Systems (IROS), 2019, pp. 4020-4026.
\newline
\newline
@INPROCEEDINGS\{eitel19iros,\\
author = \{A. \{Eitel\} and N. \{Hauff\} and W. \{Burgard\}\}, \\
booktitle = \{2019 IEEE/RSJ International Conference on Intelligent Robots and Systems (IROS)\}, \\
title = \{Self-supervised Transfer Learning for Instance Segmentation through Physical Interaction\}, \\
year = \{2019\}, \\
volume = \{\}, \\
number = \{\}, \\
pages = \{4020-4026\}, \\
\}

\newpage
\twocolumn

\maketitle
\thispagestyle{plain}
\pagestyle{plain}

\begin{abstract}
  Instance segmentation of unknown objects from images is regarded as
  relevant for several robot skills including grasping, tracking and
  object sorting.  Recent results in computer vision have shown that
  large hand-labeled datasets enable high segmentation performance. To
  overcome the time-consuming process of manually labeling data for
  new environments, we present a transfer learning approach for robots
  that learn to segment objects by interacting with their environment in
  a self-supervised manner.  Our robot pushes unknown objects on a
  table and uses information from optical flow to create training
  labels in the form of object masks.  To achieve this, we fine-tune an
  existing DeepMask network for instance segmentation on the self-labeled
  training data acquired by the robot.   We evaluate our trained
  network (SelfDeepMask) on a set of real images showing challenging
  and cluttered scenes with novel objects.  Here, SelfDeepMask
  outperforms the DeepMask network trained on the COCO dataset by
  9.5\% in average precision. Furthermore, we combine our approach with 
  recent approaches for training with noisy labels in order to
  better cope with induced label noise.

\end{abstract}

\section{INTRODUCTION}
\label{section:introduction}
The ability to segment object instances in a category-agnostic manner, i.e., to partition individual objects regardless of the
class, is necessary to enhance the visual perception capabilities of
a robot for various manipulation tasks, e.g., object instance grasping
or object sorting. Additionally, it can be useful for identifying
target objects and inferring their spatial relationships from the
visual grounding of human language
instructions~\cite{jund_optimization_2018,shridhar_interactive_2018}.  In
the past, several methods have been proposed for object segmentation
based on color, texture and 3D features. However, these are known to
over-segment multi-colored objects and under-segment objects that
are adjacent~\cite{jianbo_shi_normalized_2000,richtsfeld_segmentation_2012}.

Recent segmentation methods based on deep learning require precise
hand-labeled segmentation annotations for a large number of objects as
training data~\cite{o._pinheiro_learning_2015}. Existing large-scale
datasets from the computer-vision community consist of RGB images that
show natural scenes. As these scenes differ from the typical ones
robots encounter (e.g., regarding clutter and the frontal camera viewpoint),
a common procedure to learn object perception on a robot is
to manually label a new dataset and to fine-tune a pre-trained model
given the labeled data. To reduce the labeling effort, often bounding
box detectors are used instead of pixel-wise object segmentation
methods, because labeling boxes requires less effort compared to segmenting images.

\begin{figure}[t]
  \centering
  \includegraphics[width=0.80\columnwidth]{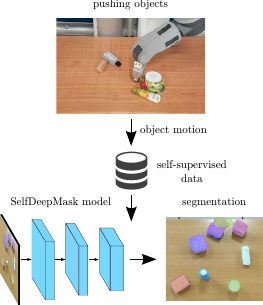}
  \caption{Our robot collects training data by interaction with its environment (top). We train a ConvNet with the automatically labeled dataset (left). Instance segmentation result using our trained model (right).\label{fig:cover}
  Implementation:~\url{https://github.com/aeitel/self_deepmask}
  }
\vspace{-4.5mm}
\end{figure}

\begin{figure*}[ht]
  \centering
  \includegraphics[width=0.99\textwidth]{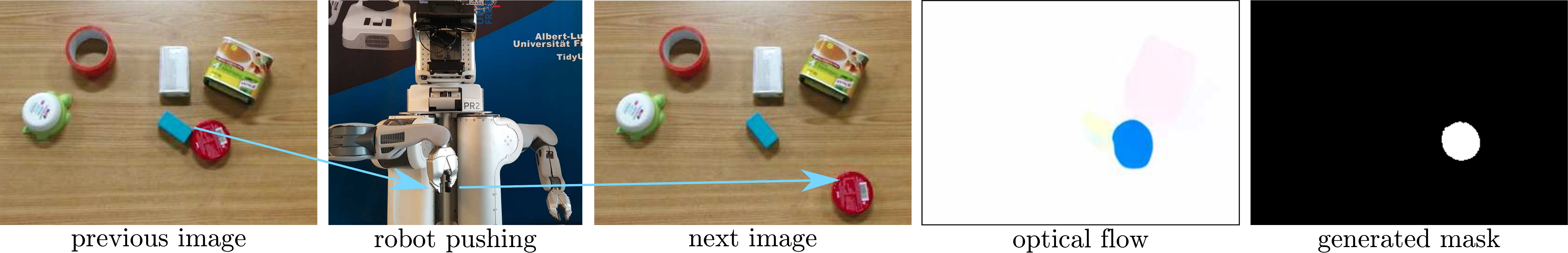}
  \caption{We use motion between two consecutive images as primary self-supervisory signal to automatically generate object masks. The robot creates object motion
   by pushing an object. We filter and cluster the resulting optical flow field into segments (not depicted here) to generate a binary mask that we add to the training set.
  \label{fig:mask_generation}
  }
\end{figure*}

Commonly, instance segmentation is considered as an offline process,
where labeling is manually performed before training and at test time
there is no mechanism to correct mistakes of the learned model. One of
the main challenges for robots is to adapt their own perceptual
capabilities to new environments.  To address this challenge, we
present a method that performs interactive data collection and uses a
self-supervised labeling process for adapting the trained model to a
novel scenario. Our robot interacts with its environment to collect
and label its own data by using manipulation primitives and by
observing the outcome of its own actions. The overall concept is
depicted in Figure~\ref{fig:cover}. It is inspired by research in interactive
perception that aims to resolve perception ambiguities via
interaction~\cite{kenney_interactive_2009,bohg_interactive_2017} and
combines it with the idea of self-supervised learning.

Our approach works as follows: The robot moves an object on the
surface of a table in front of it using push actions and collects one
RGB image before and one after each action. By detecting coherent
motion of pixels in the two images, it creates a binary object mask
that selects the moved object. We use this binary mask as a
self-supervisory signal, so that our self-supervised
method for transfer learning does not require any hand-labeled data. Following this
approach, we collect a diverse set of training data consisting of
various object instances to fine-tune DeepMask~\cite{o._pinheiro_learning_2015}, a recent instance segmentation CNN pre-trained on the COCO dataset with 886K labeled object instances~\cite{lin_microsoft_2014}. We show improved segmentation results using our transfer learning
method (SelfDeepMask) compared to the pre-trained DeepMask network.
We employ SelfDeepMask as part of an interactive object-separation experiment
and show that the segmentation performance at test time increases with
each action.

The contributions of this paper are: (1) a self-supervised method that
generates labeled training data in form of object masks acquired from
robot--object interactions, (2) a method that uses motion information
from two consecutive images in combination with the end-effector position to generate a high-quality object mask, (3) real-world
experiments evaluating the generalization abilities of our
SelfDeepMask network to segment unseen objects in clutter.

\section{RELATED WORK}
This work builds on prior research in interactive segmentation and self-supervised learning for robotics and computer vision. The main motivation is to improve perception for robot manipulation, e.g., 
grasping and placing.
\label{section:related}
\subsection{Interactive Segmentation}
Previous work in interactive segmentation considers segmenting specific scenes in an interactive manner using image differencing techniques or visual feature tracking
to update the segmentation after each action~\cite{fitzpatrick_first_2003,kenney_interactive_2009,bjorkman_active_2010,schiebener_segmentation_2011,hausman_tracking-based_2013,van_hoof_probabilistic_2014}.
Patten~\etal~\cite{patten_action_2018} extend prior methods by enabling online segmentation also during the interaction.
We use the perceptual signal from robot--object interactions to create a segmentation mask. Aforementioned online methods do not use learning to generalize to new object scenes,
whereas our approach learns a visual model that improves segmentation for new object scenes using data gathered from over 2,300 robot--object interactions.
Katz~\etal~\cite{katz_perceiving_2014} present an interactive segmentation algorithm based on a learned model for detecting favourable actions to remove object clutter.
An interesting recent overview on interactive perception is presented by Bohg~\etal~\cite{bohg_interactive_2017}, who summarize that perception is facilitated
by interaction with the environment. Inspired by this line of research we use interaction to improve perception and provide a novel perspective by using interactive perception for self-supervised learning.
To improve segmentation performance, we first collect training data through interaction that we use for self-supervised transfer learning.
To further improve segmentation performance at test time, our robot uses its ability to interact with objects (i.e., we follow the usual scheme of interactive perception) together with the transferred network for instance segmentation. We will see this example in Fig.~\ref{fig:singulation_ap}.

Closest to our work is the recent method by Pathak~\etal~\cite{pathak_learning_2018} that uses grasping for self-supervised instance segmentation. As opposed to grasping objects,
we show results for pushing, which allows to learn arbitrary objects and not only graspable ones. Furthermore, we use motion information to generate masks, while their method
uses simple frame differencing, which is less robust to movement of multiple objects as we show in our experiments.

\subsection{Self-supervised Robot Learning}
Several recent works have used self-supervised learning for acquiring manipulation skills such as grasping~\cite{pinto_supersizing_2016,murali_cassl:_2018,levine_learning_2018},
regrasping~\cite{chebotar_self-supervised_2016}, pushing~\cite{agrawal_learning_2016}, combined pushing and grasping~\cite{zeng_learning_2018}, pouring~\cite{sermanet_time-contrastive_2018} and
tool affordance understanding~\cite{mar_self-supervised_2017}. Pinto~\etal~\cite{pinto_curious_2016} learn visual feature representations from manipulation interactions. Several works use
multiple views as self-supervision signal for 6D pose estimation, which can be used complementary to our method~\cite{zeng_multi-view_2017,mitash_self-supervised_2017}.
None of the mentioned approaches leverage self-supervised learning for instance segmentation.
Pot~\etal~\cite{pot_self-supervisory_2018} learn a bounding box detector in a self-supervised manner by navigating in static environments and associating frames using
Simultaneous Localization and Mapping. Wellhausen~\etal~\cite{wellhausen_where_2019} train a terrain segmentation network in a self-supervised manner
by navigating with a legged robot.

\subsection{Self-supervised Visual Learning}
Schmidt~\etal~\cite{schmidt_self-supervised_2017} learn feature descriptors for dense correspondence.
Ovsep~\etal~\cite{osep_large-scale_2017} discover objects in street scene videos using tracking.
Pathak~\etal~\cite{pathak_learning_2017} learn to segment objects by tracking their movement, but only consider single object videos that are passively observed.
Milan~\etal~\cite{milan_semantic_2018} present semi-automatic data labeling for semantic segmentation.
Aforementioned methods are self-supervised but not interactive. More recently, Danielczuk~\etal~\cite{danielczuk_segmenting_2019} train a depth-based network for instance segmentation with rendered data from a simulator and show successful
transfer to the real world. As we use RGB data, training in simulation is not straightforward because it raises several domain-adaptation challenges,
which are not the focus of this work.

\section{SELF-SUPERVISED INSTANCE SEGMENTATION BY INTERACTION}
\label{section:method}
In this section, we describe our approach for self-supervised instance
segmentation. We require that all objects are movable and are placed on a flat surface.
Based on this, we describe how to realize the interactive data collection, how
our self-supervised mask generation works and how
we perform transfer learning with the autonomously
gathered data.

\subsection{Interactive Data}

To acquire the data necessary for adapting the model, we need a robot
that is able to push objects on a table in front of it with its
end effector. In our current approach, we sample push actions horizontally
and parallel to the table plane, which we segment using depth
information. The robot needs to be able to generate pushes that move objects
into free space and keeps them within it.  In comparison to random pushing
this mitigates the problem of moving objects into each other or moving two
objects at the same time. One approach for realizing this has been described in our
previous work~\cite{eitel_learning_2020}. We employ this method in the
approach described in this paper. Given a robust method for object pushing, human
involvement can be rather limited and is only needed to exchange the
objects on the table. In principle, this task could also be performed
automatically using systems like Dex-Net~\cite{mahler_dex-net_2017}. To
perform trajectory planning, one can use any existing approach
like the LBKPIECE motion planning algorithm provided by the Open Motion Planning Library~\cite{sucan_open_2012}, which we also utilize in
this paper. Please note that with our approach, depth information is
only needed during data collection to extract the surface of the
table. In principle the surface can also be extracted using tactile
data. Independent of this, at execution time, our method only requires
RGB image data.

In our current system, we use a PR2 robot equipped with a Kinect~2
head camera that provides RGB-D images with a resolution of
$960\times540$ pixels. We use both robot arms to enable covering the
whole workspace. The overall setup for learning is depicted in
Figure~\ref{fig:mask_generation}.

\subsection{Self-supervisory Signal}
We use coherent motion of object pixels as the primary supervision
signal, see Fig~\ref{fig:mask_generation}.  The robot captures
images $\bo_t$ and $\bo_{t+1}$ before and after each push action respectively
$\ba_t = (x_{push},y_{push})$. We represent the push action as the
pixel in the image where the push started and leverage the stored
end-effector state during the interaction together with the known
camera--robot extrinsic calibration. Before and after a push
interacion, the robot arms are positioned such that they do not
obstruct the view for capturing $\bo_t$ and $\bo_{t+1}$. The goal is
to create a labeled training dataset
$\mathcal{D} = \lbrace (\bo^{1}, \bs^{1}), \dots, (\bo^{N}, \bs^{N})
\rbrace$ that consists of images and automatically-labeled
segmentation masks $\bs^n$. However, not every $\bo_t$ is added to $\mathcal{D}$.
During data collection, we perform the following steps in each time step $t$:

We compute the optical flow $\bu_t,\bv_t = \mathit{flow}(\bo_{t},\bo_{t+1})$
using the FlowNet2 network~\cite{ilg_flownet_2017}. We filter the optical flow
field by setting all flow vectors that point to the estimated table
ground plane to zero. Second, we cluster the filtered flow
$\bu_{t}^*,\bv_{t}^*$ using normalized graph cuts
$\bS_t = \mathit{clusters}(\bu_{t}^*,\bv_{t}^*)$ to obtain a set of segments
$\bS_t = \{ \bs_{t}^1,\dots, \bs_{t}^L \}$. To handle rotating
objects we remove segments from $\bS_t$ where the corresponding flow
magnitudes on the segment exceed a standard deviation threshold (of
15.0).  Next, we pick the segmentation mask $\bs_{t}^l\in\bS_t$ containing the
push action pixel $\ba_t$. Using the push action as a prior
information enforces that only segments from $\bS_t$ are used that
overlap with the end-effector position at the beginning of the push.
Finally, we add both the image $\bo_{t}$ and the segmentation mask
$\bs_{l}$ to the training set $\mathcal{D}$.  Note that using our
approach we add at most one mask per time step $t$ to the training set
together with the associated image $\bo_{t}$.  We further discard
complete interactions where the mean magnitude of all optical-flow vectors in the image
exceeds a given threshold (of 7.0), to handle
scenes with large motions.
\begin{figure}[t]
  \centering
  \includegraphics[width=0.99\columnwidth]{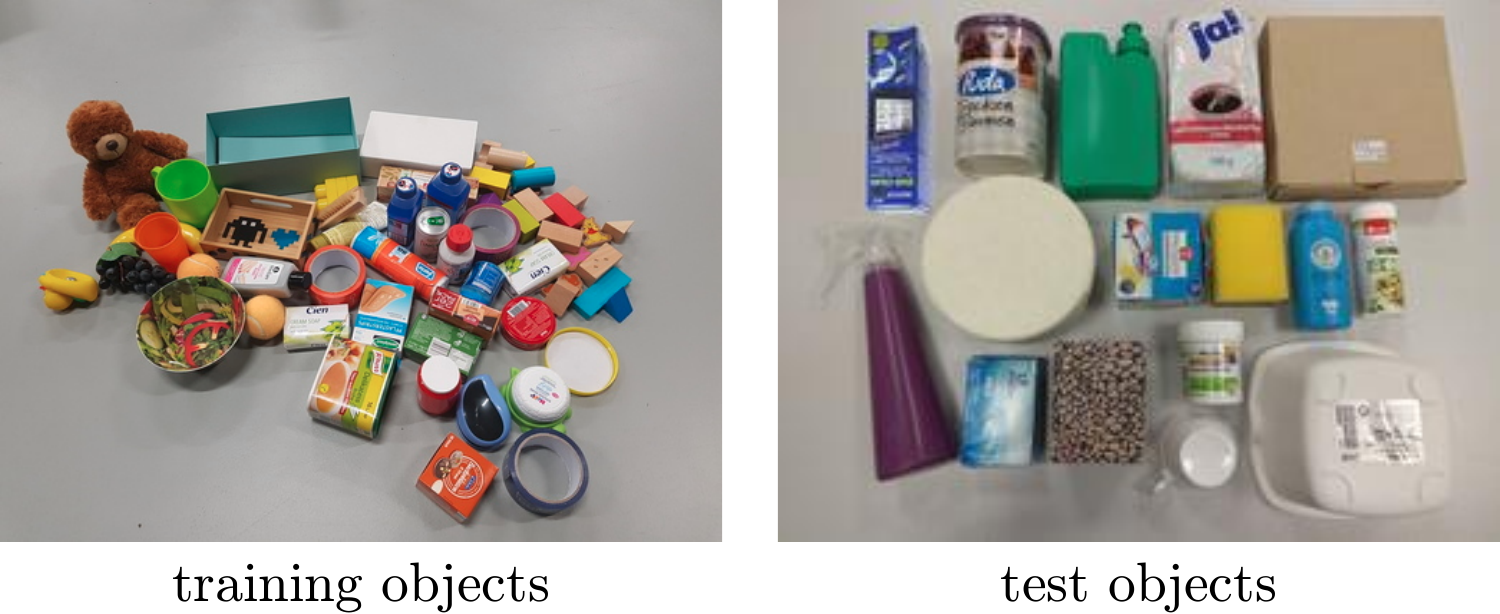}
  \caption{The set of objects used for training and testing.\label{fig:objects}}
\end{figure}

\begin{table*}[t]
    \centering
    \caption{Quantitative segmentation results}
    \label{tab:results}
    \begin{tabular}{c c c c c}
	  	Method &  Trained on & AP@0.5 & AP@0.75 & AP@0.5:0.95 \\
	  	\hline
   		\hline
	  	DeepMask   		          & COCO 	& 59.3 & 52.0 & 40.0 \\
	  	SharpMask 		          & COCO 	& 59.4 & 48.9 & 37.8 \\
	  	DeepMask with NMS 				& COCO 	& 71.4  & 58.1 & 45.9 \\
	  	SharpMask with NMS  			& COCO  & 71.6  & 54.5 & 43.6 \\
	  	DeepMask frame differencing  			& COCO \& \ourdataset 	& 71.2 $\pm$ 6.7 & 45.9 $\pm$ 8.3 & 39.6 $\pm$ 5.7 \\
	  	\hline
      Ours, SelfDeepMask, 1.3k interactions                              & COCO \& \ourdataset	& 76.8 $\pm$ 2.9 & 57.9 $\pm$ 3.5 & 47.1 $\pm$ 2.5	\\
      DeepMask, human-labeled						& COCO \& \ourdataset 	& 84.7 $\pm$ 0.9 & 66.0 $\pm$ 1.6 & 53.1 $\pm$ 0.7 \\
        \hline
      Ours, SelfDeepMask, 2.3k interactions                              & COCO \& \ourdataset	& 80.0 $\pm$ 1.5 & 57.1 $\pm$ 1.3 &  47.5 $\pm$ 1.3 \\
      Ours, SelfDeepMask, 2.3k interactions, co-teaching~\cite{NIPS2018_8072} & COCO \& \ourdataset & 80.9 $\pm$ 0.7 & 64.8 $\pm$ 0.4 & 51.9 $\pm$ 0.6 \\

    \end{tabular}
\end{table*}

\subsection{Network Transfer Learning}
\label{section:method_transfer}
We use a state-of-the-art method for category-agnostic instance
segmentation known as DeepMask~\cite{o._pinheiro_learning_2015}.  The core
of DeepMask is a ConvNet, which jointly predicts a mask and an object
score for an image patch. At test time, the network is fed with RGB
image patches in a sliding-window manner using multiple scales. If the
score network detects that the center pixel of the image patch belongs
to an object it triggers the mask network to produce a corresponding
mask. We use our interactive method to fine-tune the score network.
The usage of DeepMask is complementary to our contribution of learning
in a self-supervised manner from robot--object interactions, i.e., a
different segmentation network could also be used.

Given $\bo^{n}$ and $\bs^{n}$ sampled from $\mathcal{D}$ we follow the
default image preprocessing steps of DeepMask.  The image is resized
to different scales (from $2^2$ to $2^1$ with a step of $2^{1/2}$) and
into an image size of $224\times224$.  The score network of
DeepMask is trained by sampling positive and negative image patches.
DeepMask considers an image patch as positive if it contains an object in a canonical
position in the middle of the patch. To account for noise, we jitter
positive examples in translation (of $\pm 16$ pixels) and scale
deformation (of $2^{\pm1/4}$). We label an image patch as a negative
example if it is at least $\pm32$ pixels or $2^1$ in scale away from a
canonical positive example. We enhance the data augmentation pipeline of DeepMask, which by default consists of vertical image flipping ($p = 0.25$) with a $(0-360\degree)$ rotation ($p = 0.25$) of both images and our
automatically labeled segmentation masks, which increases robustness to rotations.

All scenes contain multiple objects but our method only labels one object per image.
The training loss can be large if the pre-trained model assigns a high confidence to a true positive object
in the image that is missing a label. To handle this issue we use bootstrapping, i.e., we use the predictions
of the pre-trained network to relabel these potentially false labels~\cite{szegedy_scalable_2014}.
In practice we set the gradient to zero for image patches, for which the pre-trained model
assigns a confidence greater than 0.5 for class ``object'' while the label denotes class ``background''.



We fine-tune our SelfDeepMask network for ten epochs with a learning
rate of 0.001, using stochastic gradient descent with momentum of 0.9 and
weight decay of 0.0005. We train each model with five different random seeds
to report an uncertainty measure of the final performance and to take into account
that we are training with noisy training data.
Furthermore, we add non-maximum suppression (NMS) at test time to both DeepMask and our
SelfDeepMask to remove false overlapping detections.

\section{EXPERIMENTS}
\label{section:experiments}
In this section, we evaluate the instance segmentation performance of
our approach on the \ourdataset dataset.  \ourdataset contains 2,300
training, 50 validation and 190 test scenes (images) of diverse
real-world objects. The training set contains over 50 different
objects and the test set 16 novel objects to examine generalization
performance, see Fig~\ref{fig:objects}. The test scenes contain
between six and eight objects, with various levels of clutter. We manually
annotated masks in the validation and test scenes for evaluation.  We compare
against two state-of-the-art category-agnostic methods for instance segmentation
called DeepMask~\cite{o._pinheiro_learning_2015} and
SharpMask~\cite{pinheiro_learning_2016}, both trained with over 886K
labeled object instances from the COCO dataset. To improve the
performance of the two baselines we add NMS and filter out large
masks that cannot correspond to objects in the \ourdataset
dataset. We also implement a frame-differencing baseline similar to Pathak~\etal~\cite{pathak_learning_2018},
in which we replaced the optical-flow-based mask generation, while keeping the rest of our method fixed.
We train two variants of our SelfDeepMask with different amounts of training data: one trained with
800 training images gathered from 1.3k interactions and one trained with 1.5k images gathered from
2.3k interactions. 
Finally, we report results in which a human labels 300 training images in a pixel-wise manner.

All data in \ourdataset is collected autonomously by the robot, which uses a learning-based method for object separation (from own prior work)  that effectively isolates cluttered
objects using push actions~\cite{eitel_learning_2020}.
In a second experiment, we perform a fine-grained evaluation of segmentation
performance with respect to each push interaction, which provides insights in
segmentation performance for various degrees of clutter.
We use the same 190 images for evaluation.
The NMS threshold is optimized on the validation set. We found a value of 0.5
for SharpMask and DeepMask to give best results. For our SelfDeepMask
we set the NMS threshold to 0.4.

\begin{table*}[t]
    \centering
    \caption{Ablation studies}
    \label{tab:ablation}
    \begin{tabular}{c c c c }
	  	Method & AP@0.5 & AP@0.75 & AP@0.5:0.95 \\
	  	\hline
   		\hline
      	SelfDeepMask, 1.3k interactions            	& 76.8 $\pm$ 2.9 & 57.9 $\pm$ 3.5 & 47.1 $\pm$ 2.5\\
	  	\hline
   		SelfDeepMask without action as prior         & 73.7 $\pm$ 2.7 & 54.9 $\pm$ 3.4 & 45.5 $\pm$ 2.3 \\
   		SelfDeepMask without bootstrapping		       & 74.6 $\pm$ 1.3 & 54.7 $\pm$ 2.0 & 45.1 $\pm$ 1.4 \\
   		SelfDeepMask without bootstrapping, fine-tune mask head & 71.9 $\pm$ 1.5 & 45.1 $\pm$ 1.8 & 38.9 $\pm$ 0.5 \\
   		SelfDeepMask without table filtering	       & 66.5 $\pm$ 3.3 & 29.2 $\pm$ 5.8 & 31.8 $\pm$ 2.8 \\
    \end{tabular}
\end{table*}
 
\begin{table*}[t]
    \centering
    \caption{Experiments with methods for combating label noise and training data from 2.3k interactions.}
    \label{tab:label_noise}
    \begin{tabular}{c c c c }
	  	Method & AP@0.5 & AP@0.75 & AP@0.5:0.95 \\
	  	\hline
   		\hline
      	SelfDeepMask, 2.3k interactions, no noisy-label filtering            	& 72.9 $\pm$ 1.5 & 50.0 $\pm$ 2.0 & 42.6 $\pm$ 1.4\\
	  	\hline
   		SelfDeepMask, reed-hard~\cite{reed2014training}         & 76.3 $\pm$ 0.8 & 53.5 $\pm$ 0.8 & 45.3 $\pm$ 1.0 \\
        SelfDeepMask, bootstrapping heuristic                              & 80.0 $\pm$ 1.5 & 57.1 $\pm$ 1.3 &  47.5 $\pm$ 1.3 \\
   		SelfDeepMask, self-paced~\cite{NIPS2010_3923}        & 80.4 $\pm$ 0.1 & 64.7 $\pm$ 0.2 & 51.0 $\pm$ 0.1 \\
   		SelfDeepMask, small-loss sampling~\cite{NIPS2018_8072} & 80.7 $\pm$ 0.7 & 64.4 $\pm$ 0.7 & 51.5 $\pm$ 0.4 \\
   		SelfDeepMask, co-teaching~\cite{NIPS2018_8072} & 80.9 $\pm$ 0.7 & 64.8 $\pm$ 0.4 & 51.9 $\pm$ 0.6 \\
    \end{tabular}
\end{table*}

\subsection{Quantitative Comparisons}
We compare the performance of the methods using the standard COCO
instance segmentation benchmark metric. The metric that we report is
average precision (AP) over different IoU (intersection over union) thresholds (AP from 0.5 to
0.95, AP at 0.5 and AP at 0.75). Higher AP indicates better
performance and higher IoU thresholds penalize localization errors of
the methods. The results are shown in Table~\ref{tab:results}. Our
SelfDeepMask outperforms SharpMask with NMS and also improves the AP performance for two of the three
IoU thresholds with respect to DeepMask with NMS. The results indicate that
our self-supervised transfer learning approach is able to further
improve the performance of a system that is already trained on large
amounts of labeled data.
Our results further show that using motion information from optical flow
outperforms generating training masks based on frame differencing.
Moreover, almost doubling the amount of data results in a moderate improvement
in performance and reduces the variance.
In addition, combining our approach with Co-teaching~\cite{NIPS2018_8072}, a recent method for coping with noisy labels, further improves the performance.


\subsection{Ablation Studies}
We conduct several experiments in which we remove one step of our method, see Table~\ref{tab:ablation}.
We observe that the final performance is lower if the action information is not used as a prior to select the mask.
Similarly, turning off bootstrapping reduces performance because our method does not
account for true positives generated by the pre-trained network in combination with missing object annotations in our training set.
Furthermore, we find that fine-tuning the mask head in addition to the score head of DeepMask deteriorates performance, which shows that
the mask head is more sensitive to training with noisy masks.
Finally, skipping the step of filtering out motion of pixels that map to the table also reduces performance.
\begin{figure}[t]
\centering
\includegraphics[width=0.99\columnwidth]{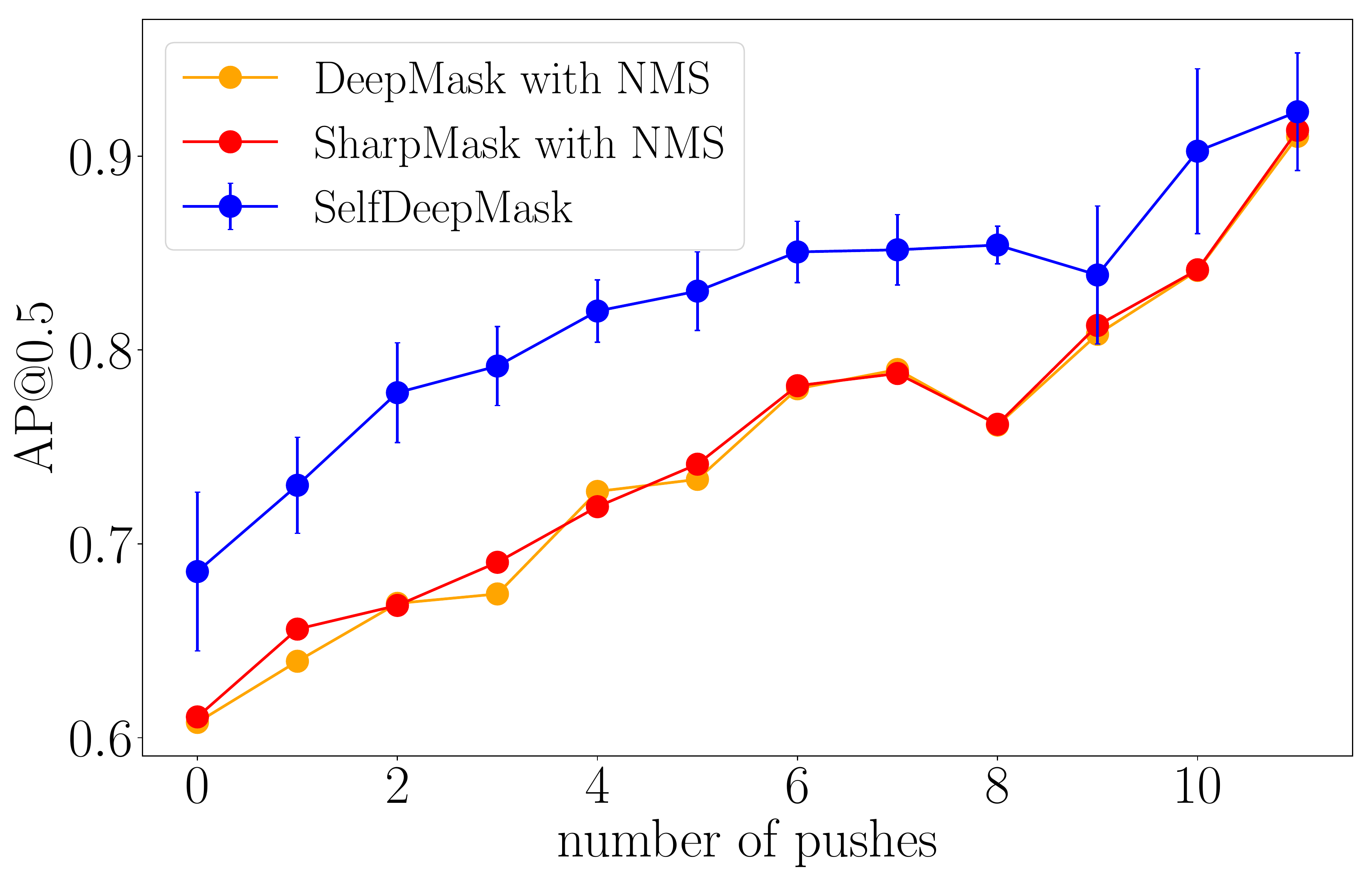}
\caption{Interactive segmentation experiment. The robot separates
  cluttered objects using pushing, which increases the segmentation
  performance of our SelfDeepMask network at test time after each
  push. SelfDeepMask achieves higher performance in cluttered scenes, showing higher average precision when then
  robot has performed only few or no pushes. Results are averaged over 5 models, trained with different random seeds. Error bars denote the standard deviation
  of the average precision.
  We use the model trained with 2.3k interactions.}
\label{fig:singulation_ap}       
\end{figure}

\subsection{Learning with Noisy Labels}
Real-world data, annotated in a self-supervised manner can be noisy. This is especially the case for a robot that collects and labels its own data as in our case.
In this section we test various existing methods that we combine with our SelfDeepMask network to combat label noise. In most existing methods, sample reweighting or removing high-loss samples are commonly used strategies to cope with noisy labels.
We implemented five recent methods that modify the classification loss in the score network and compare them in Table~\ref{tab:label_noise}.
\\ 
\textbf{Hard bootstrap~\cite{reed2014training}}: The loss term consists of a convex weighted combination of predicted and original labels. We set the weighting factor $\beta$ to 0.7.\\
\textbf{Self-paced learning~\cite{NIPS2010_3923}}: Self-paced training uses a predefined curriculum, which skips certain data points that are considered to be yet too hard (measured by the per-example loss). It uses a threshold $\lambda$ to distinguish between easy/hard examples that is increased every epoch with a growing factor (of 1.2). We start with $\lambda = 0.002$, which corresponds to the average loss in the first epoch (a value we took from prior experiments).\\
\textbf{Small-loss sampling~\cite{NIPS2018_8072}}: This approach is similar to self-paced learning but removes a fixed amount of samples in each batch. We found that removing $10\%$ of samples sorted based on high loss values yields best results.\\
\textbf{Co-teaching~\cite{NIPS2018_8072}}: It uses the same small-loss sampling scheme but adds a second network. In each mini-batch of data, each network views its small-loss instances and selects the useful instances for its peer network to update the parameters.\\
\textbf{Bootstrapping heuristic}: Our initial bootstrapping approach that we described in section~\ref{section:method_transfer}.\\
For all methods we tested the last and the best (using early-stopping) model snapshot and report the higher numbers only. 
\begin{figure*}[t]
\centering
\includegraphics[width=0.90\textwidth]{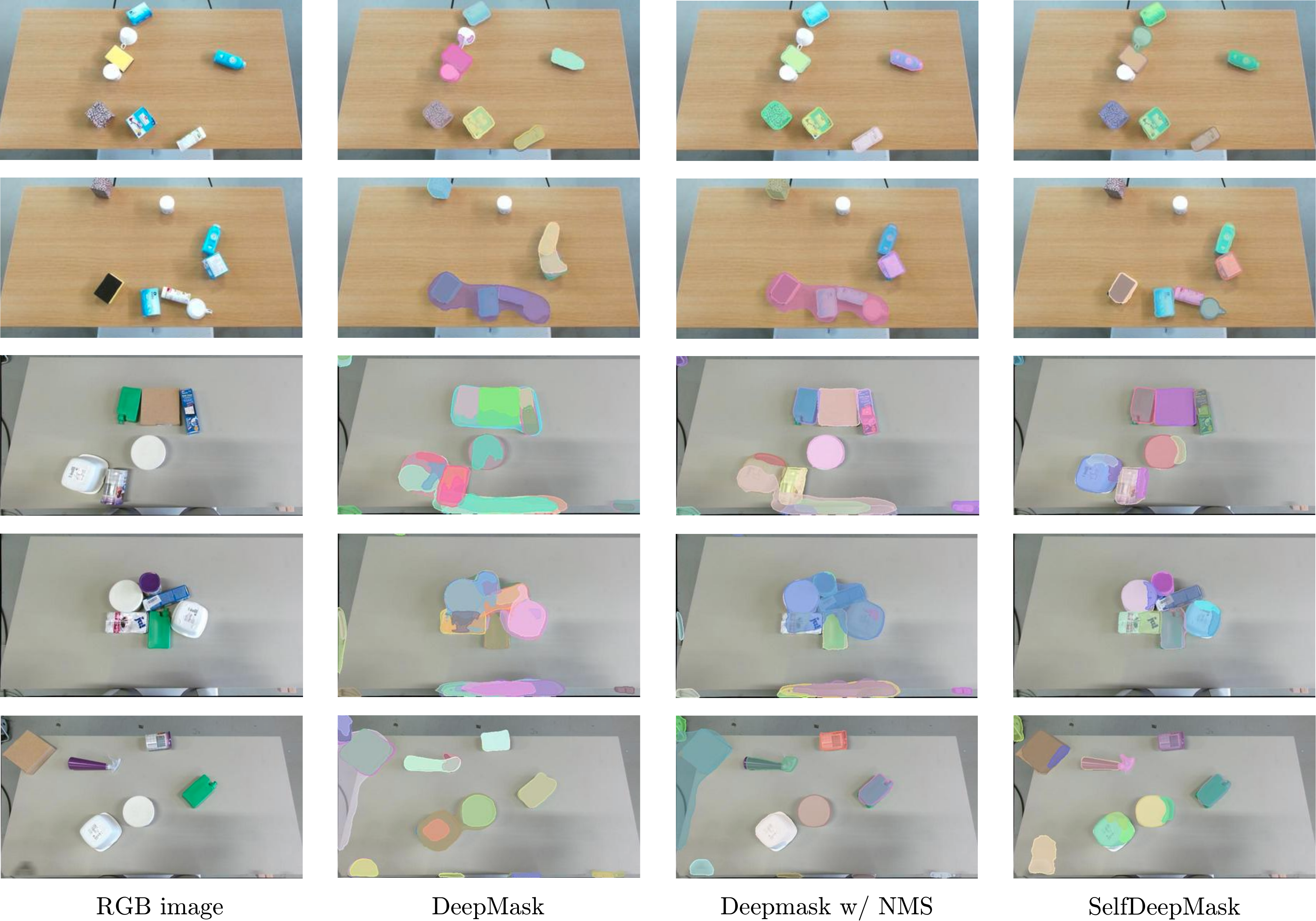}
\caption{Visualization of predicted instance segmentation masks generated by DeepMask (left), DeepMask with NMS (middle) and our SelfDeepMask trained with 1.3k interactions (right). The top three rows show examples
where our method predicts accurate masks. The fourth row shows that all methods produce failures for very cluttered scenes. The last row shows that DeepMask produces more false positives at
the borders of the workspace.}
\label{fig:qualitative_results}       
\end{figure*}
\begin{figure*}[t]
\centering
\includegraphics[width=0.90\textwidth]{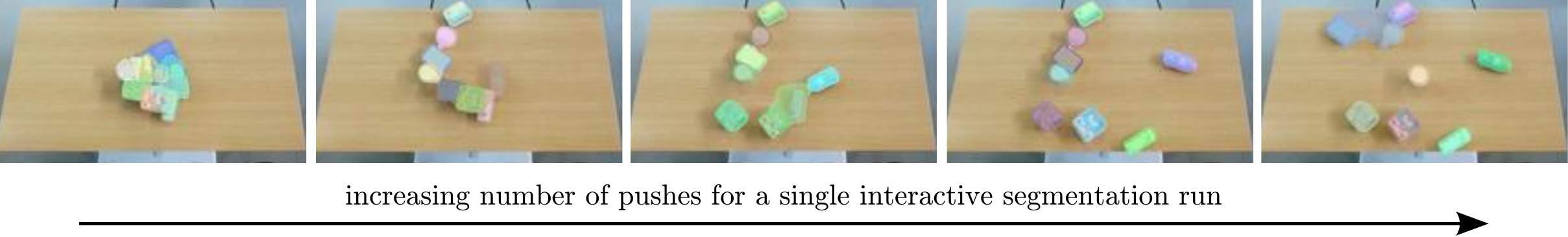}
\caption{Exemplary trial of the interactive segmentation experiment. Segmentation accuracy improves with each interaction due to minimization of clutter, see video at~\url{https://bit.ly/38WXKlx}.}
\label{fig:singulation_experiment}       
\end{figure*}

\subsection{Interactive Instance Segmentation with Object Separation}
In this second experiment we show that combining the paradigm of interactive
segmentation with object separation improves overall perception performance at test time. Figure~\ref{fig:singulation_ap} shows that the segmentation performance improves with each
interaction up to 23.7 average precision points compared to a passive
segmentation where the initial scene is not changed by the
robot. In total the robot interactively segmented 20 scenes consisting of unknown objects.
Figure~\ref{fig:singulation_experiment} qualitatively shows the
improved instance segmentation after each push interaction. The push actions
were chosen based on a pre-trained CNN for object separation from own prior work~\cite{eitel_learning_2020}.
The results suggest that following an interactive perception strategy substantially improves the segmentation performance at test time and in addition provides the self-supervised data for training a more accurate network.

\subsection{Qualitative Results}
Figure~\ref{fig:qualitative_results} shows the outputs generated by the
different methods.  Our method generalizes and segments novel objects
effectively. Qualitatively, our method performs similarly to DeepMask.
All methods produce failures for highly
cluttered scenes but our method is less prone to clutter.

\section{CONCLUSIONS}
\label{section:conclusions}
In this work, we presented a self-supervised transfer learning
approach for instance segmentation that leverages physical robot
interaction with its environment to automatically generate a training
dataset for adapting pre-trained networks to the current
environment. Instead of labeling object masks in an expensive manual
procedure, our robot learns to generate object masks by observing the
outcome of its own interaction with objects. As the main supervision
signal we use motion information from pushing objects.  Our results
suggest that fine-tuning a pre-trained model with the automatically
labeled data substantially improves the segmentation performance. The
more high-level take-home message from this is that robots can in fact
improve their perception performance, achieved by manually
labeled large-scale datasets, by physically interacting with their
environment. 
We also showed that we can further improve the performance of our method
if we leverage recent algorithms for training with noisy labels.
In future work, we will
explore how to fine-tune our model in a new environment without
forgetting the previously learned knowledge, particularly when the
adaptation is carried out for longer periods of time. 

\addtolength{\textheight}{-0.0cm}   

\section*{Acknowledgments}
This work was partly funded by the German Research Foundation (DFG)
under the priority program Autonomous Learning SPP 1527, under grant number EXC1086
as well as by the Federal Ministry of Education and Research (BMBF) under Deep PTL and OML.
We thank Oier Mees for his help. We also thank the anonymous reviewers for suggestions.


\bibliographystyle{IEEEtran}
\bibliography{iros19}

\begin{thebibliography}{10}
\providecommand{\url}[1]{#1}
\csname url@samestyle\endcsname
\providecommand{\newblock}{\relax}
\providecommand{\bibinfo}[2]{#2}
\providecommand{\BIBentrySTDinterwordspacing}{\spaceskip=0pt\relax}
\providecommand{\BIBentryALTinterwordstretchfactor}{4}
\providecommand{\BIBentryALTinterwordspacing}{\spaceskip=\fontdimen2\font plus
\BIBentryALTinterwordstretchfactor\fontdimen3\font minus
  \fontdimen4\font\relax}
\providecommand{\BIBforeignlanguage}[2]{{%
\expandafter\ifx\csname l@#1\endcsname\relax
\typeout{** WARNING: IEEEtran.bst: No hyphenation pattern has been}%
\typeout{** loaded for the language `#1'. Using the pattern for}%
\typeout{** the default language instead.}%
\else
\language=\csname l@#1\endcsname
\fi
#2}}
\providecommand{\BIBdecl}{\relax}
\BIBdecl

\bibitem{jund_optimization_2018}
P.~Jund, A.~Eitel, N.~Abdo, and W.~Burgard, ``Optimization {Beyond} the
  {Convolution}: {Generalizing} {Spatial} {Relations} with {End}-to-{End}
  {Metric} {Learning},'' in \emph{2018 {IEEE} {International} {Conference} on
  {Robotics} and {Automation} ({ICRA})}, May 2018, pp. 4510--4516, iSSN:
  2577-087X.

\bibitem{shridhar_interactive_2018}
\BIBentryALTinterwordspacing
M.~Shridhar and D.~Hsu, ``Interactive {Visual} {Grounding} of {Referring}
  {Expressions} for {Human}-{Robot} {Interaction},'' in \emph{Robotics:
  {Science} and {Systems} {XIV}}, vol.~14, Jun. 2018. [Online]. Available:
  \url{http://www.roboticsproceedings.org/rss14/p28.html}
\BIBentrySTDinterwordspacing

\bibitem{jianbo_shi_normalized_2000}
J.~Shi and J.~Malik, ``Normalized cuts and image segmentation,'' \emph{IEEE
  Transactions on Pattern Analysis and Machine Intelligence}, vol.~22, no.~8,
  pp. 888--905, Aug. 2000.

\bibitem{richtsfeld_segmentation_2012}
A.~Richtsfeld, T.~Mörwald, J.~Prankl, M.~Zillich, and M.~Vincze,
  ``Segmentation of unknown objects in indoor environments,'' in \emph{2012
  {IEEE}/{RSJ} {International} {Conference} on {Intelligent} {Robots} and
  {Systems}}, Oct. 2012, pp. 4791--4796, iSSN: 2153-0858.

\bibitem{o._pinheiro_learning_2015}
\BIBentryALTinterwordspacing
P.~O. O.~Pinheiro, R.~Collobert, and P.~Dollar, ``Learning to {Segment}
  {Object} {Candidates},'' in \emph{Advances in {Neural} {Information}
  {Processing} {Systems} 28}, C.~Cortes, N.~D. Lawrence, D.~D. Lee,
  M.~Sugiyama, and R.~Garnett, Eds.\hskip 1em plus 0.5em minus 0.4em\relax
  Curran Associates, Inc., 2015, pp. 1990--1998. [Online]. Available:
  \url{http://papers.nips.cc/paper/5852-learning-to-segment-object-candidates.pdf}
\BIBentrySTDinterwordspacing

\bibitem{kenney_interactive_2009}
J.~Kenney, T.~Buckley, and O.~Brock, ``Interactive segmentation for
  manipulation in unstructured environments,'' in \emph{2009 {IEEE}
  {International} {Conference} on {Robotics} and {Automation}}, May 2009, pp.
  1377--1382, iSSN: 1050-4729.

\bibitem{bohg_interactive_2017}
J.~Bohg, K.~Hausman, B.~Sankaran, O.~Brock, D.~Kragic, S.~Schaal, and G.~S.
  Sukhatme, ``Interactive {Perception}: {Leveraging} {Action} in {Perception}
  and {Perception} in {Action},'' \emph{IEEE Transactions on Robotics},
  vol.~33, no.~6, pp. 1273--1291, Dec. 2017.

\bibitem{lin_microsoft_2014}
T.-Y. Lin, M.~Maire, S.~Belongie, J.~Hays, P.~Perona, D.~Ramanan, P.~Dollár,
  and C.~L. Zitnick, ``\BIBforeignlanguage{en}{Microsoft {COCO}: {Common}
  {Objects} in {Context}},'' in \emph{\BIBforeignlanguage{en}{Computer {Vision}
  – {ECCV} 2014}}, ser. Lecture {Notes} in {Computer} {Science}, D.~Fleet,
  T.~Pajdla, B.~Schiele, and T.~Tuytelaars, Eds.\hskip 1em plus 0.5em minus
  0.4em\relax Cham: Springer International Publishing, 2014, pp. 740--755.

\bibitem{fitzpatrick_first_2003}
P.~Fitzpatrick, ``First contact: an active vision approach to segmentation,''
  in \emph{Proceedings 2003 {IEEE}/{RSJ} {International} {Conference} on
  {Intelligent} {Robots} and {Systems} ({IROS} 2003) ({Cat}. {No}.03CH37453)},
  vol.~3, Oct. 2003, pp. 2161--2166 vol.3.

\bibitem{bjorkman_active_2010}
M.~Björkman and D.~Kragic, ``Active 3d scene segmentation and detection of
  unknown objects,'' in \emph{2010 {IEEE} {International} {Conference} on
  {Robotics} and {Automation}}, May 2010, pp. 3114--3120, iSSN: 1050-4729.

\bibitem{schiebener_segmentation_2011}
D.~Schiebener, A.~Ude, J.~Morimoto, T.~Asfour, and R.~Dillmann, ``Segmentation
  and learning of unknown objects through physical interaction,'' in \emph{2011
  11th {IEEE}-{RAS} {International} {Conference} on {Humanoid} {Robots}}, Oct.
  2011, pp. 500--506, iSSN: 2164-0572.

\bibitem{hausman_tracking-based_2013}
K.~Hausman, F.~Balint-Benczedi, D.~Pangercic, Z.-C. Marton, R.~Ueda, K.~Okada,
  and M.~Beetz, ``Tracking-based interactive segmentation of textureless
  objects,'' in \emph{2013 {IEEE} {International} {Conference} on {Robotics}
  and {Automation}}, May 2013, pp. 1122--1129, iSSN: 1050-4729.

\bibitem{van_hoof_probabilistic_2014}
H.~van Hoof, O.~Kroemer, and J.~Peters, ``Probabilistic {Segmentation} and
  {Targeted} {Exploration} of {Objects} in {Cluttered} {Environments},''
  \emph{IEEE Transactions on Robotics}, vol.~30, no.~5, pp. 1198--1209, Oct.
  2014.

\bibitem{patten_action_2018}
T.~Patten, M.~Zillich, and M.~Vincze, ``Action {Selection} for {Interactive}
  {Object} {Segmentation} in {Clutter},'' in \emph{2018 {IEEE}/{RSJ}
  {International} {Conference} on {Intelligent} {Robots} and {Systems}
  ({IROS})}, Oct. 2018, pp. 6297--6304, iSSN: 2153-0858.

\bibitem{katz_perceiving_2014}
\BIBentryALTinterwordspacing
D.~Katz, A.~Venkatraman, M.~Kazemi, J.~A. Bagnell, and A.~Stentz,
  ``\BIBforeignlanguage{en}{Perceiving, learning, and exploiting object
  affordances for autonomous pile manipulation},''
  \emph{\BIBforeignlanguage{en}{Autonomous Robots}}, vol.~37, no.~4, pp.
  369--382, Dec. 2014. [Online]. Available:
  \url{https://doi.org/10.1007/s10514-014-9407-y}
\BIBentrySTDinterwordspacing

\bibitem{pathak_learning_2018}
D.~Pathak, Y.~Shentu, D.~Chen, P.~Agrawal, T.~Darrell, S.~Levine, and J.~Malik,
  ``Learning {Instance} {Segmentation} by {Interaction},'' in \emph{2018
  {IEEE}/{CVF} {Conference} on {Computer} {Vision} and {Pattern} {Recognition}
  {Workshops} ({CVPRW})}, Jun. 2018, pp. 2123--21\,233, iSSN: 2160-7508.

\bibitem{pinto_supersizing_2016}
L.~Pinto and A.~Gupta, ``Supersizing self-supervision: {Learning} to grasp from
  50k tries and 700 robot hours,'' in \emph{2016 {IEEE} {International}
  {Conference} on {Robotics} and {Automation} ({ICRA})}, May 2016, pp.
  3406--3413.

\bibitem{murali_cassl:_2018}
A.~Murali, L.~Pinto, D.~Gandhi, and A.~Gupta, ``{CASSL}: {Curriculum}
  {Accelerated} {Self}-{Supervised} {Learning},'' in \emph{2018 {IEEE}
  {International} {Conference} on {Robotics} and {Automation} ({ICRA})}, May
  2018, pp. 6453--6460, iSSN: 2577-087X.

\bibitem{levine_learning_2018}
\BIBentryALTinterwordspacing
S.~Levine, P.~Pastor, A.~Krizhevsky, J.~Ibarz, and D.~Quillen,
  ``\BIBforeignlanguage{en}{Learning hand-eye coordination for robotic grasping
  with deep learning and large-scale data collection},''
  \emph{\BIBforeignlanguage{en}{The International Journal of Robotics
  Research}}, vol.~37, no. 4-5, pp. 421--436, Apr. 2018. [Online]. Available:
  \url{https://doi.org/10.1177/0278364917710318}
\BIBentrySTDinterwordspacing

\bibitem{chebotar_self-supervised_2016}
Y.~Chebotar, K.~Hausman, Z.~Su, G.~S. Sukhatme, and S.~Schaal,
  ``Self-supervised regrasping using spatio-temporal tactile features and
  reinforcement learning,'' in \emph{2016 {IEEE}/{RSJ} {International}
  {Conference} on {Intelligent} {Robots} and {Systems} ({IROS})}, Oct. 2016,
  pp. 1960--1966, iSSN: 2153-0866.

\bibitem{agrawal_learning_2016}
P.~Agrawal, A.~V. Nair, P.~Abbeel, J.~Malik, and S.~Levine, ``Learning to
  {Poke} by {Poking}: {Experiential} {Learning} of {Intuitive} {Physics},'' in
  \emph{Advances in {Neural} {Information} {Processing} {Systems} 29}, D.~D.
  Lee, M.~Sugiyama, U.~V. Luxburg, I.~Guyon, and R.~Garnett, Eds.\hskip 1em
  plus 0.5em minus 0.4em\relax Curran Associates, Inc., 2016, pp. 5074--5082.

\bibitem{zeng_learning_2018}
A.~Zeng, S.~Song, S.~Welker, J.~Lee, A.~Rodriguez, and T.~Funkhouser,
  ``Learning {Synergies} {Between} {Pushing} and {Grasping} with
  {Self}-{Supervised} {Deep} {Reinforcement} {Learning},'' in \emph{2018
  {IEEE}/{RSJ} {International} {Conference} on {Intelligent} {Robots} and
  {Systems} ({IROS})}, Oct. 2018, pp. 4238--4245, iSSN: 2153-0858.

\bibitem{sermanet_time-contrastive_2018}
P.~Sermanet, C.~Lynch, Y.~Chebotar, J.~Hsu, E.~Jang, S.~Schaal, S.~Levine, and
  G.~Brain, ``Time-{Contrastive} {Networks}: {Self}-{Supervised} {Learning}
  from {Video},'' in \emph{2018 {IEEE} {International} {Conference} on
  {Robotics} and {Automation} ({ICRA})}, May 2018, pp. 1134--1141, iSSN:
  2577-087X.

\bibitem{mar_self-supervised_2017}
T.~Mar, V.~Tikhanoff, G.~Metta, and L.~Natale, ``Self-supervised learning of
  tool affordances from 3d tool representation through parallel {SOM}
  mapping,'' in \emph{2017 {IEEE} {International} {Conference} on {Robotics}
  and {Automation} ({ICRA})}, May 2017, pp. 894--901.

\bibitem{pinto_curious_2016}
L.~Pinto, D.~Gandhi, Y.~Han, Y.-L. Park, and A.~Gupta,
  ``\BIBforeignlanguage{en}{The {Curious} {Robot}: {Learning} {Visual}
  {Representations} via {Physical} {Interactions}},'' in
  \emph{\BIBforeignlanguage{en}{Computer {Vision} – {ECCV} 2016}}, ser.
  Lecture {Notes} in {Computer} {Science}, B.~Leibe, J.~Matas, N.~Sebe, and
  M.~Welling, Eds.\hskip 1em plus 0.5em minus 0.4em\relax Cham: Springer
  International Publishing, 2016, pp. 3--18.

\bibitem{zeng_multi-view_2017}
A.~Zeng, K.-T. Yu, S.~Song, D.~Suo, E.~Walker, A.~Rodriguez, and J.~Xiao,
  ``Multi-view self-supervised deep learning for 6d pose estimation in the
  {Amazon} {Picking} {Challenge},'' in \emph{2017 {IEEE} {International}
  {Conference} on {Robotics} and {Automation} ({ICRA})}, May 2017, pp.
  1386--1383.

\bibitem{mitash_self-supervised_2017}
C.~Mitash, K.~E. Bekris, and A.~Boularias, ``A self-supervised learning system
  for object detection using physics simulation and multi-view pose
  estimation,'' in \emph{2017 {IEEE}/{RSJ} {International} {Conference} on
  {Intelligent} {Robots} and {Systems} ({IROS})}, Sep. 2017, pp. 545--551,
  iSSN: 2153-0866.

\bibitem{pot_self-supervisory_2018}
\BIBentryALTinterwordspacing
E.~Pot, A.~Toshev, and J.~Kosecka, ``Self-supervisory {Signals} for {Object}
  {Discovery} and {Detection},'' \emph{CoRR}, vol. abs/1806.03370, 2018.
  [Online]. Available: \url{http://arxiv.org/abs/1806.03370}
\BIBentrySTDinterwordspacing

\bibitem{wellhausen_where_2019}
L.~Wellhausen, A.~Dosovitskiy, R.~Ranftl, K.~Walas, C.~Cadena, and M.~Hutter,
  ``Where {Should} {I} {Walk}? {Predicting} {Terrain} {Properties} {From}
  {Images} {Via} {Self}-{Supervised} {Learning},'' \emph{IEEE Robotics and
  Automation Letters}, vol.~4, no.~2, pp. 1509--1516, Apr. 2019.

\bibitem{schmidt_self-supervised_2017}
T.~Schmidt, R.~Newcombe, and D.~Fox, ``Self-{Supervised} {Visual} {Descriptor}
  {Learning} for {Dense} {Correspondence},'' \emph{IEEE Robotics and Automation
  Letters}, vol.~2, no.~2, pp. 420--427, Apr. 2017.

\bibitem{osep_large-scale_2017}
\BIBentryALTinterwordspacing
A.~Osep, P.~Voigtlaender, J.~Luiten, S.~Breuers, and B.~Leibe, ``Large-{Scale}
  {Object} {Discovery} and {Detector} {Adaptation} from {Unlabeled} {Video},''
  \emph{CoRR}, vol. abs/1712.08832, 2017. [Online]. Available:
  \url{http://arxiv.org/abs/1712.08832}
\BIBentrySTDinterwordspacing

\bibitem{pathak_learning_2017}
D.~Pathak, R.~Girshick, P.~Dollár, T.~Darrell, and B.~Hariharan, ``Learning
  {Features} by {Watching} {Objects} {Move},'' in \emph{2017 {IEEE}
  {Conference} on {Computer} {Vision} and {Pattern} {Recognition} ({CVPR})},
  Jul. 2017, pp. 6024--6033, iSSN: 1063-6919.

\bibitem{milan_semantic_2018}
A.~Milan, T.~Pham, K.~Vijay, D.~Morrison, A.~Tow, L.~Liu, J.~Erskine,
  R.~Grinover, A.~Gurman, T.~Hunn, N.~Kelly-Boxall, D.~Lee, M.~McTaggart,
  G.~Rallos, A.~Razjigaev, T.~Rowntree, T.~Shen, R.~Smith, S.~Wade-McCue,
  Z.~Zhuang, C.~Lehnert, G.~Lin, I.~Reid, P.~Corke, and J.~Leitner, ``Semantic
  {Segmentation} from {Limited} {Training} {Data},'' in \emph{2018 {IEEE}
  {International} {Conference} on {Robotics} and {Automation} ({ICRA})}, May
  2018, pp. 1908--1915, iSSN: 2577-087X.

\bibitem{danielczuk_segmenting_2019}
M.~Danielczuk, M.~Matl, S.~Gupta, A.~Li, A.~Lee, J.~Mahler, and K.~Goldberg,
  ``Segmenting {Unknown} 3d {Objects} from {Real} {Depth} {Images} using {Mask}
  {R}-{CNN} {Trained} on {Synthetic} {Data},'' in \emph{2019 {International}
  {Conference} on {Robotics} and {Automation} ({ICRA})}, May 2019, pp.
  7283--7290, iSSN: 1050-4729.

\bibitem{eitel_learning_2020}
A.~Eitel, N.~Hauff, and W.~Burgard, ``Learning to {Singulate} {Objects} {Using}
  a {Push} {Proposal} {Network},'' in \emph{Robotics {Research}}, N.~M. Amato,
  G.~Hager, S.~Thomas, and M.~Torres-Torriti, Eds.\hskip 1em plus 0.5em minus
  0.4em\relax Cham: Springer International Publishing, 2020, pp. 405--419.

\bibitem{mahler_dex-net_2017}
\BIBentryALTinterwordspacing
J.~Mahler, J.~Liang, S.~Niyaz, M.~Laskey, R.~Doan, X.~Liu, J.~Aparicio, and
  K.~Goldberg, ``Dex-{Net} 2.0: {Deep} {Learning} to {Plan} {Robust} {Grasps}
  with {Synthetic} {Point} {Clouds} and {Analytic} {Grasp} {Metrics},'' in
  \emph{Robotics: {Science} and {Systems} {XIII}}, vol.~13, Jul. 2017.
  [Online]. Available: \url{http://www.roboticsproceedings.org/rss13/p58.html}
\BIBentrySTDinterwordspacing

\bibitem{sucan_open_2012}
I.~A. Sucan, M.~Moll, and L.~E. Kavraki, ``The {Open} {Motion} {Planning}
  {Library},'' \emph{IEEE Robotics Automation Magazine}, vol.~19, no.~4, pp.
  72--82, Dec. 2012.

\bibitem{ilg_flownet_2017}
E.~Ilg, N.~Mayer, T.~Saikia, M.~Keuper, A.~Dosovitskiy, and T.~Brox,
  ``{FlowNet} 2.0: {Evolution} of {Optical} {Flow} {Estimation} with {Deep}
  {Networks},'' in \emph{2017 {IEEE} {Conference} on {Computer} {Vision} and
  {Pattern} {Recognition} ({CVPR})}, Jul. 2017, pp. 1647--1655, iSSN:
  1063-6919.

\bibitem{NIPS2018_8072}
B.~Han, Q.~Yao, X.~Yu, G.~Niu, M.~Xu, W.~Hu, I.~Tsang, and M.~Sugiyama,
  ``Co-teaching: Robust training of deep neural networks with extremely noisy
  labels,'' in \emph{Advances in Neural Information Processing Systems 31},
  S.~Bengio, H.~Wallach, H.~Larochelle, K.~Grauman, N.~Cesa-Bianchi, and
  R.~Garnett, Eds.\hskip 1em plus 0.5em minus 0.4em\relax Curran Associates,
  Inc., 2018, pp. 8527--8537.

\bibitem{szegedy_scalable_2014}
\BIBentryALTinterwordspacing
C.~Szegedy, S.~E. Reed, D.~Erhan, and D.~Anguelov, ``Scalable, {High}-{Quality}
  {Object} {Detection},'' \emph{CoRR}, vol. abs/1412.1441, 2014. [Online].
  Available: \url{http://arxiv.org/abs/1412.1441}
\BIBentrySTDinterwordspacing

\bibitem{pinheiro_learning_2016}
P.~O. Pinheiro, T.-Y. Lin, R.~Collobert, and P.~Dollár,
  ``\BIBforeignlanguage{en}{Learning to {Refine} {Object} {Segments}},'' in
  \emph{\BIBforeignlanguage{en}{Computer {Vision} - {ECCV} 2016}}, ser. Lecture
  {Notes} in {Computer} {Science}, B.~Leibe, J.~Matas, N.~Sebe, and M.~Welling,
  Eds.\hskip 1em plus 0.5em minus 0.4em\relax Cham: Springer International
  Publishing, 2016, pp. 75--91.

\bibitem{reed2014training}
\BIBentryALTinterwordspacing
S.~E. Reed, H.~Lee, D.~Anguelov, C.~Szegedy, D.~Erhan, and A.~Rabinovich,
  ``Training deep neural networks on noisy labels with bootstrapping,'' in
  \emph{3rd International Conference on Learning Representations, {ICLR} 2015,
  San Diego, CA, USA, May 7-9, 2015, Workshop Track Proceedings}, Y.~Bengio and
  Y.~LeCun, Eds., 2015. [Online]. Available:
  \url{http://arxiv.org/abs/1412.6596}
\BIBentrySTDinterwordspacing

\bibitem{NIPS2010_3923}
\BIBentryALTinterwordspacing
M.~P. Kumar, B.~Packer, and D.~Koller, ``Self-paced learning for latent
  variable models,'' in \emph{Advances in Neural Information Processing Systems
  23}, J.~D. Lafferty, C.~K.~I. Williams, J.~Shawe-Taylor, R.~S. Zemel, and
  A.~Culotta, Eds.\hskip 1em plus 0.5em minus 0.4em\relax Curran Associates,
  Inc., 2010, pp. 1189--1197. [Online]. Available:
  \url{http://papers.nips.cc/paper/3923-self-paced-learning-for-latent-variable-models.pdf}
\BIBentrySTDinterwordspacing

\end{thebibliography}

\end{document}